# Fast Machine Learning Method with Vector Embedding on Orthonormal Basis and Spectral Transform

Louis Yu Lu

E-mail: louisyulu@gmail.com

## Abstract

This paper presents a novel fast machine learning method that leverages two techniques: Vector Embedding on Orthonormal Basis (VEOB) and Spectral Transform (ST). The VEOB converts the original data encoding into a vector embedding with coordinates projected onto orthonormal bases. The Singular Value Decomposition (SVD) technique is used to calculate the vector basis and projection coordinates, leading to an enhanced distance measurement in the embedding space and facilitating data compression by preserving the projection vectors associated with the largest singular values. On the other hand, ST transforms sequence of vector data into spectral space. By applying the Discrete Cosine Transform (DCT) and selecting the most significant components, it streamlines the handling of lengthy vector sequences. The paper provides examples of word embedding, text chunk embedding, and image embedding, implemented in Julia language with a vector database. It also investigates unsupervised learning and supervised learning using this method, along with strategies for handling large data volumes.

## 1. Introduction

First, we highlight the issues in popular ML techniques such as deep learning and transformer model: the slow learning pace with gradient descends, and the extensive memory consumption of the attention mechanism in transformer for long vector sequences.

Next, we introduce VEOB and particularly the SVD technique for dimension reduction and data compression. ST and specifically Discrete Cosine Transform (DCT) are introduced to tackle the memory consumption issue of long sequence embedding. A data partition and clustering algorithm is also presented for processing the vector samples.

Several examples are implemented to illustrate the application of VEOB and ST:

- Word orthography/spelling embedding using SVD and DCT.
- Word semantic/meaning with SVD.
- Word semantic/meaning embedding combining orthography and dictionary entries with synonyms and antonyms.
- Sentence embedding with word vectors and DCT.
- MNIST data with 2D DCT and followed by SVD for dimension reduction.

We further explore both unsupervised and supervised learning techniques using this approach, as well as strategies for handling large volumes of data. Finally, we draw conclusions and propose potential areas for future investigation.

## 2. Problems and strategy for the solution

We aim to address the following issues:

- Common deep learning and transformer techniques use back propagation on gradient descending algorithms, which are notably slow, necessitate substantial computing resources, and are generally costly for model training.
- The vectors in the intermediate layers of these methods are hardly interpretable, they lack orthogonality, which can distort the distance measures for similarity comparison, and it's even more challenging to reduce the dimension without losing significant latent information encoded in the vector components.
- The memory consumption increases rapidly in the order of $O(n^2)$ with the transformer's attention mechanism as the number $n$ of the vectors in the sequence grows.

The strategy for addressing these issues is as follows:
- Find a fast forward method to generate vectors embedding the original data with reduced dimensions, lower the necessary computing resource and save on model training expense.
- Ensure that embedding vectors are positioned on an orthonormal basis, which allows for a more precise comparison of similarity on Euclidean distance and reduction of vector dimension by discarding less important components.
- Use spectral transform to manage the vector sequence length and consequently decrease memory usage.

## 3. SVD method, orthonormal basis, and projection vectors for embedding

The widely recognized Singular Value Decomposition (SVD) technique from linear algebra is employed to determine the orthonormal basis and the projection vectors on this basis.

Given $n$ vector $x_i$ with dimension $m$, an $m \times n$ matrix can be formed:

$$X = [x_1 \ x_2 \ ... \ x_n]$$

Suppose $m \leq n$, $X$ can be decomposed into three separate matrices:

$$X = U \ \Sigma \ V^T$$

Where $U$ is a $m \times m$ matrix with $m$ orthonormal column vectors, each of dimension $m$.

$\Sigma$ is a diagonal matrix with $m$ singular values, if $X$ only contains real values, theses singular values $\sigma_i$ are either positive or zero.

$V^T$ is the transpose of $n \times m$ matrix $V$, which contains $m$ orthonormal column vectors of dimension $n$.

The column vectors of $U$ are selected as the orthonormal basis, the projection of $X$ on the basis is computed as follows:

$$P = U^T X = \Sigma \ V^T$$

$P$ consists of $n$ vectors of dimension $m$:

$$P = [p_1 \ p_2 \ ... \ p_n]$$

Using popular SVD algorithms, the largest elements in $\Sigma$ are arranged in descending order from the top, corresponding to the column vectors $U$ on the left side. Given an integer $k < m$, $X$ can be approximated as follows:

$$X \approx \widetilde{U} \ \widetilde{\Sigma} \ \widetilde{V}^T$$

Where $\widetilde{U}$ is a $m \times k$ submatrix of $U$ and contains $k$ orthonormal vectors of dimension $m$.

$\widetilde{\Sigma}$ is a diagonal submatrix containing first $k$ values from $\Sigma$.

$\widetilde{V}^T$ is the transpose of $n \times k$ submatrix $\widetilde{V}$ from $V$ and contains $k$ orthonormal vectors of dimension $n$.

The projection of $X$ on the dimension reduced basis is computed as follows:

$$\widetilde{P} = \widetilde{U}^T X = \widetilde{\Sigma} \, \widetilde{V}^T$$

$\widetilde{P}$ consists of $n$ column vectors of dimension $k$, which is selected as the embedding vectors:

$$\widetilde{P} = [p_1 \; p_2 \; \ldots \; p_n]$$

Each vector $p_i$ has lower dimension than the original encoding vector $x_i$ but still closely represents the original data's distribution.

As $p_i$ lies on the orthonormal basis, Euclidian distance measurement is more effective for similarity comparison.

The prerequisites for applying SVD to the source matrix $X$ are relatively flexible; it doesn't necessitate symmetric or semi-positive conditions as in the case of principal component analysis. This flexibility simplifies the construction of the source matrix for practical uses, examples of which will be demonstrated in the following sections.

## 4. Spectral transform and discrete cosine transform

The vector sequence can be interpreted as a trajectory in a high-dimensional space, which can be transformed into spectral space. In our scenarios, the vectors are discontinuous from one point to another along the path, hence we opted for the Discrete Cosine Transform (DCT) due to its simplicity among the Fourier transform family.

Given a sequence of integers $k$ with $N$ data points, there exists $N$ orthogonal cosine functions:

$$s(N, m, k) = \cos\left(\frac{\pi}{N-1}(N-m)(k-1)\right) \; for \; 1 \le m \le N, 1 \le k \le N$$

These constitute a comprehensive series of spectral functions. The original vector data series or embedding vectors series can be transformed into the spectral space using the following formula:

$$f_m = \sum_{k=1}^{N} s(N, m, k) x_k \quad for \; 1 \le m \le N$$

Where $f_m$ and $x_k$ are vectors of identical dimension.

With word embedding and text chunk embedding, the short span of letters in a word (or words in a text chunk) are more crucial, so we can select a smaller number than the full-length $N$. The short span corresponds to high frequency in spectral analysis terminology, essentially applying a high pass filter. $f_m$ addresses the question of whether a vector of length $m$ exists in the full sequence, but not exactly where - this is the desired position agnostic feature required in vector embedding applications.

## 5. Data partition into clustered cells

In the following examples, we utilize a data partition algorithm to process the data. This algorithm partitions a given set of vector samples and clusters them into Voronoi cells. Each of these cells contains a centroid, a maximum radius, and a list of samples that it encompasses.

The partitioning process is achieved by a hyperplane that intersects the centroid of the samples. The longest axis of the data distribution, which is the normal of this hyperplane, effectively divides the data along the direction of maximum variance. This axis is also determined using an efficient random SVD method.

The partitioned subsamples can be further divided using the same method in a recursive manner until either the specified number of partition levels is reached, or the resulting subsamples count falls below a certain number.

Throughout this process, a partition tree is constructed. The centroids of the leaves are used as the cell centroids. The time complexity of this process is $O(n \log(n))$.

A separate scan is then required to allocate the samples to their respective cells. Each sample is assigned to a cell with the closest centroid. Given that the number of cells is on the order of $\log(n)$, this scan of $n$ samples also has a time complexity of $O(n \log(n))$. Therefore, the overall time complexity of the entire partition and clustering algorithm is $O(n \log(n))$.

## 6.  Implementation Examples

The source code for the examples can be found at this GitHub repository:
https://github.com/louisyulu/veob-and-st

### 6.1 Sample data sources, development tools and hardware configuration

The examples utilize publicly available open-source datasets:

- IMDB Large Movie Review Dataset https://ai.stanford.edu/~amaas/data/sentiment/ with 50,000 text files
- A simple English dictionary https://github.com/nightblade9/simple-english-dictionary with data in json format
- MNIST digits and fashion images training and testing samples downloaded from https://app.activeloop.ai/public/

The Julia language is employed to leverage its performance and math libraries, with both Julia and Python packages used for implementation.

Julia packages:

- Pluto, PlutoUI, HypertextLiteral and Plots for the interactive notebook and UI interface
- LinearAlgebra, RandomizedLinAlg, SparseArrays and Statistics for matrix and vector calculation
- PythonCall to invoke the python functions, in the NLP preprocessing logic.

Python packages:

- Spacy for NLP processing
- ChromaBD to save the vectors data locally and perform kNN (k Nearest Neighbors) querying.

Implementation and test runs are conducted on a moderately powered laptop without GPU support:

> Apple M1 Pro Laptop
> 16 GB memory
> macOS Ventura 13.5.2

All the execution times in the following examples are measured on this computer.

### 6.2 Word orthography/spelling embedding using SVD and DCT

With 83,590 tokens derived from IMDB files, each token is encoded as a sequence of letter vectors, with each letter utilizing one hot encoding with its ASCII number to set the 1 among 127 positions, while all other positions are zeros. By only considering sub word up to 6 letters, this variable sequence of letter vectors (dimension 127) is then transformed into a fixed number of 6 vectors (dimension 127) with DCT using the parameters $N = 15, 1 \leq m \leq 6$. $N$ is twice the token's average length

By concatenating these 6 vectors into one long column vectors and combining 83,590 tokens to construct a matrix, then applying SVD to the matrix and selecting the top 40 components, an embedding vector set is achieved. The top 40 components are selected based on their significance in the singular values diagram below.

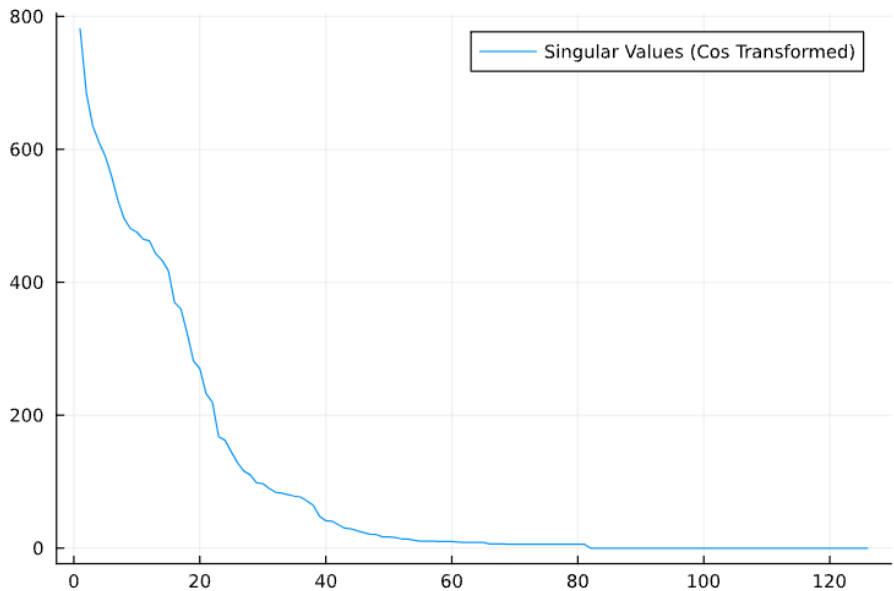

The processing time is approximately **2 seconds**.

Here are some results in orthography/spelling similarity (smaller number indicates closer similarity):

| twighlight (input) | frowning (input) |
|---|---|
| twighlight - 0.0 | frowning - 0.0 |
| twisting - 3.3822982e-6 | fronting - 1.3071899e-6 |
| twiggy - 3.3908689e-6 | frog - 1.3691999e-6 |
| twigged - 3.5892863e-6 | froggy - 1.4565417e-6 |
| twig - 4.3338027e-6 | frogs - 1.6778553e-6 |
| twigs - 4.642331e-6 | frothing - 2.3373084e-6 |
| twiggles - 4.933348e-6 | frosting - 2.3374646e-6 |
| twirling - 5.0474678e-6 | frown - 2.4049093e-6 |
| twistingly - 5.062511e-6 | frowns - 2.6923094e-6 |
| twitch - 6.360612e-6 | frowned - 2.8906518e-6 |

### 6.3 Word semantic/meaning with SVD

With 83,590 tokens and 528,460 sentences derived from IMDB files, a full scan is conducted. For each token, the context words of two words preceding and two following are examined. The column word vector is constructed with the frequency of context words before and after the current one and forms a vector of dimension 2x83,590. Then these 83,590 vectors are combined and SVD is applied, with the following diagram displaying the singular values.

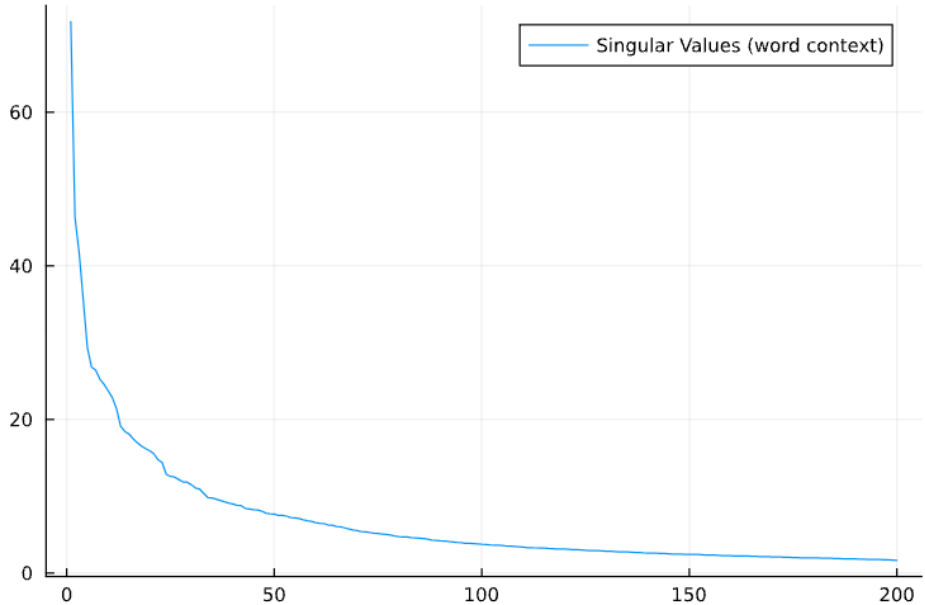

The top 50 components of the projection vectors are selected as the embedding vectors. The process takes approximately **14.6 seconds**.

Other methods to calculate the context word frequency can also be considered, such as treating the four context words as in one single bag and calculating the frequency in the bag or treating the four context positions as four separate bags and calculating the frequency in them. We choose the two-bag approach of before and after positions as it strikes a balance between accuracy and tolerance. Here are some results in semantic/meaning similarity (smaller number indicates closer similarity):

| physicist | twists |
|---|---|
| physicist - 0.0 | twists - 0.0 |
| schoolteacher - 0.020590613 | lawyers - 0.009032602 |
| reader - 0.029172964 | planes - 0.010917113 |
| correspondence - 0.030516949 | businessmen - 0.012805708 |
| gasps - 0.030873906 | bells - 0.014761503 |
| reception - 0.031360097 | sweetness - 0.014962247 |
| procedural - 0.032097798 | blades - 0.014970788 |
| secretary - 0.032877453 | intrigues - 0.016480677 |
| devious - 0.032885075 | tortures - 0.016826117 |
| lens - 0.03417209 | jews - 0.018220086 |

We noted that the outcome is more about one word substituting another in the trained sentences than the true semantics.

### 6.4 Word semantic/meaning embedding combining orthography and dictionary entries with synonyms and antonyms

The hypothesis is that dictionaries are superior sources for semantic vector embedding of words, given that humans frequently refer to dictionaries for word meanings. Therefore, we attempt to perform semantic embedding by integrating information about each word's orthographic encoding, as well as its synonyms and antonyms.

The dataset from a basic English dictionary comprises 121,340-word entries, 147,202 synonyms (the number is larger than the word entries due to the presence of multiple words), and 6,199 antonyms. The column vector for each word entry consists of three slices:

1) A dense orthographic vector of dimension 30, derived similarly to the method in 6.2 with $N = 15, 1 \leq m \leq 6$. $N$ is twice the average length of a word. The vector is normalized by dividing by the norm of singular values to prevent spelling from overly influencing the final embedding. This step takes approximately **3.3 seconds**.

2) A one-hot encoding for synonyms is used, which occupies the subsequent 147,202 slots. Only the slots corresponding to a synonym for the word entry are marked as one; all others are set to zero.

3) A one-hot antonym encoding that fills the final 6,199 slots. Only the slot for an existing antonym is marked as one; all others are zero.

By concatenating the 121,340-word column vectors, a large sparse matrix is constructed. Applying SVD to this matrix and selecting the top 300 dimensions for embedding, the following is a plot of the top 300 singular values.

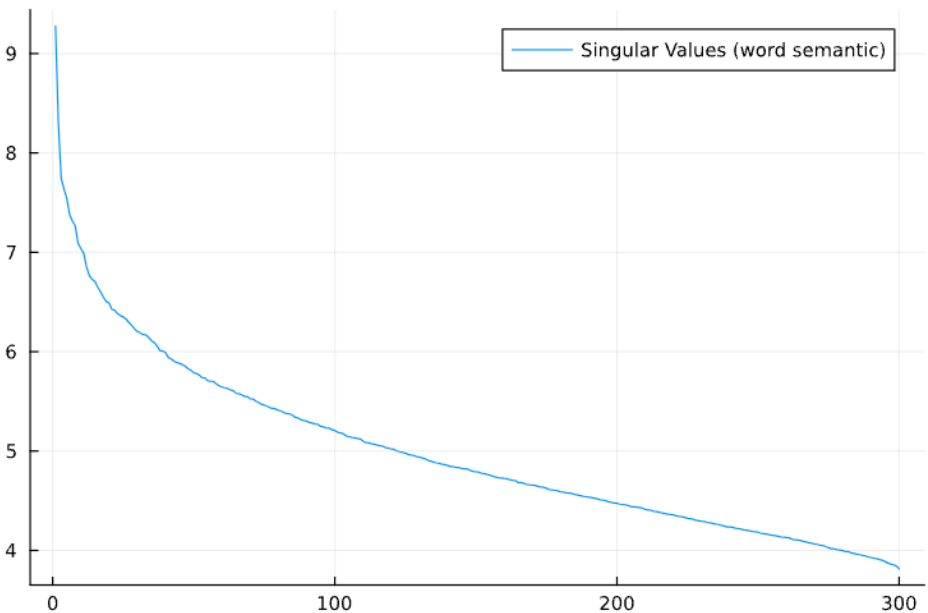

The entire process takes approximately **15.1 seconds**.

Here are some results showing semantic/meaning similarity (a smaller number indicates closer similarity):

| bus (input) | happiness (input) |
|---|---|
| bus 0.0 | happiness 0.0 |
| buses 0.0065672416 | felicity 0.009182828 |
| charabanc 0.018942868 | felicities 0.013546261 |
| motorcoach 0.018942999 | dizzied 0.038097393 |
| double-decker 0.018943142 | dizzies 0.038107295 |
| omnibus 0.018943531 | dizzying 0.03847787 |
| jitney 0.020661581 | dignifies 0.051144343 |
| charabancs 0.020742513 | broached 0.060142282 |
| autobusses 0.0207546 | authenticity 0.062201094 |
| motorbuses 0.020999735 | lavalliere 0.06267107 |

It's observed that words with similar meanings also have some spelling similarity.

## 6.5 Sentence embedding with word vectors and DCT

We select a subset of sentences from the IMDB dataset, specifically those with a maximum of 32 words. This results in a total of 528,460 sentences. Each word in the sentence is encoded using the word embedding vector from section 6.3, and DCT is applied to the sequence of word vectors with $N = 15, 1 \leq m \leq 6$, forming the sentence column vectors. SVD is then applied to the sentence matrix composed of 528,460 column vectors. The singular values are depicted in the subsequent diagram:

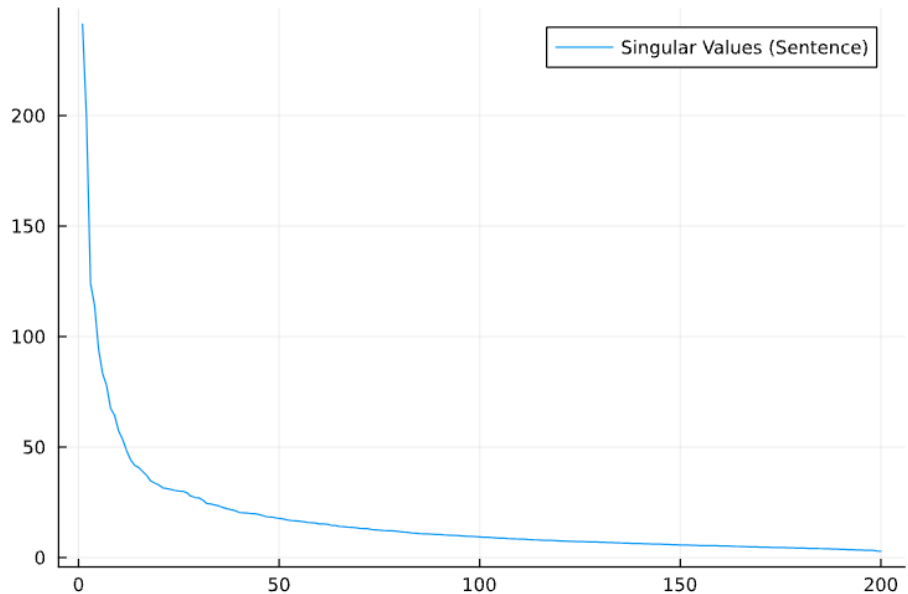

We have selected 200 as the dimension for the embedding vectors, and the processing time amounted to approximately **1.28 minutes**.

Here are some results for sentence matching, including the similarity distance and the source of each sentence:

| |
|---|
| i did like the lead actors, but thought little of the film (input) |
| >> i did like the lead actors,  but thought little of the film.  0.0 - 0_0.txt |
| >> too bad,  this could have been a thought - provoking film.    3.3087225e-24 - 10388_0.txt |
| >> what i 'd like to know is what people thought of that waiter.    3.3087225e-24 - 12136_0.txt |
| >> i suppose the pre-production was not well thought.  3.3087225e-24 - 12930_0.txt |
| >> it was the complete opposite of what i was thought it was originally.   3.3087225e-24 - 13585_0.txt |

> > considering the age of this film, i thought the dvd transfer quality excellent, with vivid colors.  3.3087225e-24 - 1545_0.txt
> > when i was 11 or 12, i thought that this was the coolest movie ever made.  3.3087225e-24 - 15514_0.txt

---

everybody has seen ' back to the future ' right?

> > everybody has seen ' back to the future ' right?  0.0 - 10001_0.txt
> > the film is set in the near future and charlton heston seems to be the only person left on the planet.  8.271806e-25 - 13023_0.txt
> > henry fonda shows a hint of his future greatness in a fabulous portrayal of julie 's no - nonsense beau.  8.271806e-25 - 1610_0.txt
> > this young woman definitely has a bright future in films.  8.271806e-25 - 16520_0.txt
> > this movie also paved the way for future stars particularly richard dreyfuss, ron howard, and harrison ford.  8.271806e-25 - 17193_0.txt
> > this series deserves to be seen by future generations. 8.271806e-25 - 18327_0.txt
> > anyone know about the possibility of a future release on vhs or dvd?  8.271806e-25 - 18404_0.txt

## 6.6 MNIST data dimension reduction with 2D DCT and SVD

Both the MNIST digit and fashion datasets consist of 60,000 training sets and 10,000 testing sets of 28x28 images. The images are initially transformed using a 2D DCT. Subsequently, SVD is applied, and the first 30 components are selected as the embedding vectors.

The singular values for the MNIST digit dataset are illustrated in the following diagram:

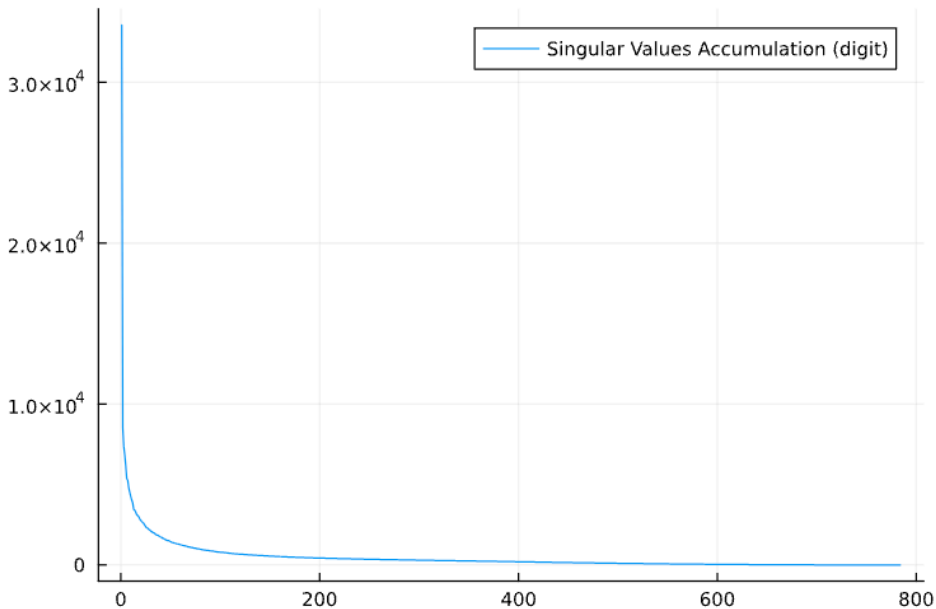

The processing time for this operation is approximately **3.6 minutes**.
Here is an example of a correct match:

Original: 9

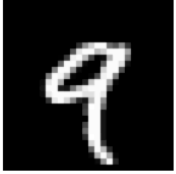

9 - 9

Matches:

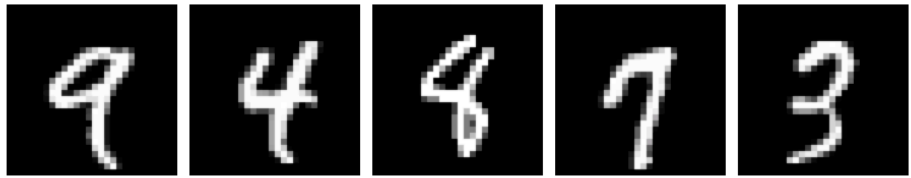

9 - 0.0          4 - 3453.3213     8 - 3603.0142     7 - 4177.9165     3 - 5780.569

And here is an example of a mismatch:

Original: 4

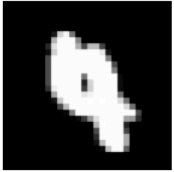

4 - 4

Matches:

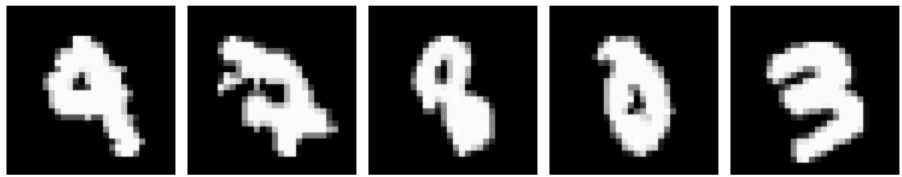

9 - 2426.1428    7 - 5506.9277     8 - 6345.3364     0 - 6880.8374     3 - 8216.376

Upon examining the top three matches, the accuracy rate is 100% for the training set and 99.4% for the testing set.

The singular values for the MNIST fashion dataset are depicted in the following diagram:

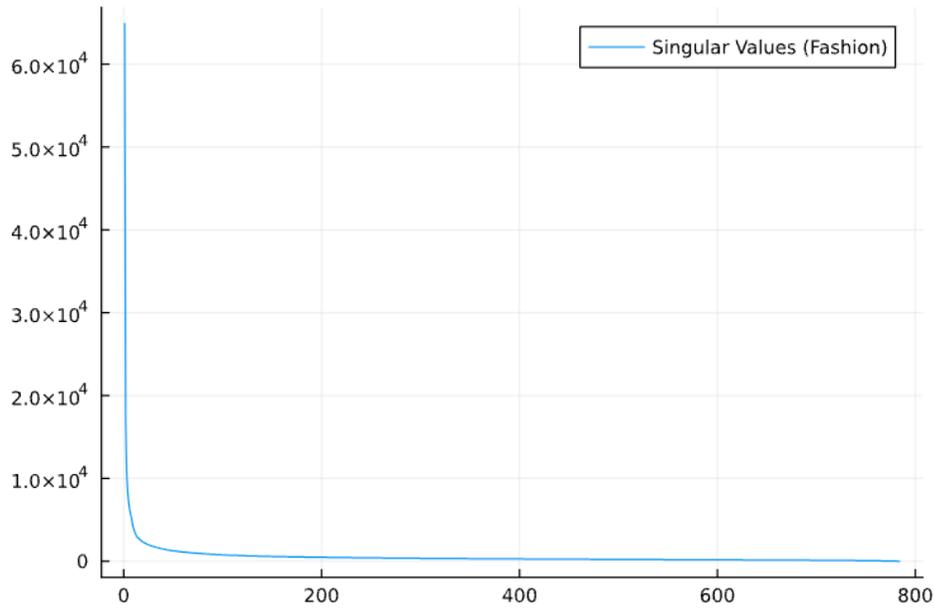

The processing time for this operation is approximately **3.8 minutes**.
Here is an example of a correct match:

Original: 9

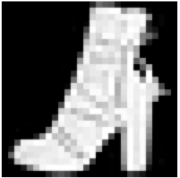

9 - Ankle-boot

Matches:

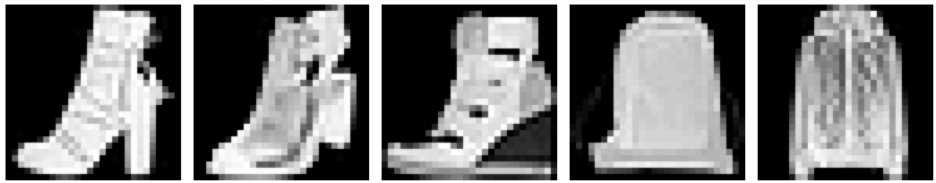

9 - 0.0    5 - 5227.483    7 - 8078.805    8 - 12124.379    4 - 13574.313

And here is an example of a mismatch:

Original: 8

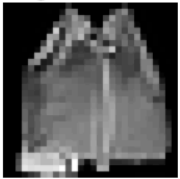

8 - Bag

Matches:

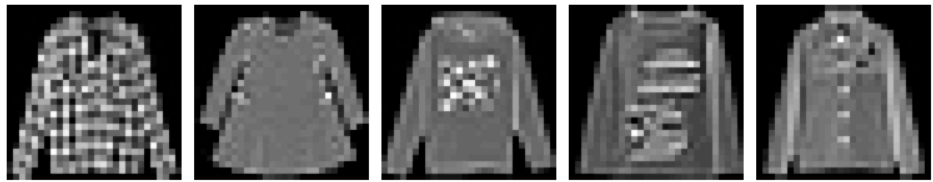

6 - 1860.5585      3 - 1896.6482      2 - 2289.8418      0 - 2375.108      4 - 2380.0771

Upon examining the top three matches, the accuracy rate is 100% for the training set and 97.7% for the testing set.

It's observed that on the testing datasets, the digit set has a higher correct matching rate than the fashion set. However, some mismatches are even challenging for humans to distinguish.

## 7. Apply VEOB and ST to unsupervised learning

In unsupervised learning, data points that exhibit similar patterns are autonomously clustered together without the need for human supervision. This process involves two crucial steps: initially, the raw source data is encoded to capture the inherent characteristics of the dataset, and subsequently, the dataset is segmented into clusters exhibiting similar patterns for querying.

### 7.1. Data encoding

There are multiple scenarios for the raw source data, and the following strategies are applied accordingly:

1) When the source data has a fixed number of attributes, which can be either categorical or numerical in nature, different encoding methods are used. For categorical data, one-hot encoding is employed, while numerical data typically requires weighting across all attributes to avoid skewing the data distribution. Each data point is then transformed into a column vector, and all vectors are combined to form a matrix. The previously described VEOB method is then applied to obtain a vector embedding. The word semantic embedding in sections 6.3 and 6.4 are as examples of this approach.

2) When the source data consists of a sequence of vectors, the ST method is applied to convert the sequence into spectral space. During this transformation, significant data sequence spans relevant to the problem are selected. Short span corresponds to high-pass filter in spectral analysis terminology, long span to low-pass filter, and specific span to band-pass filter. The transformed data with the chosen spans result in a fixed number of vectors for each data sequence, which can then be further processed with VEOB for dimension reduction. The word orthographic embedding in section 6.2 and sentence encoding in section 6.5 are examples of this approach.

3) Two-dimensional data sequences, the source 2D data is applied ST with 2D DCT, then applied VEOB for data embedding, the MNIST examples in 6.6 are examples of this approach. This approach can even be generated into 3D data or even higher dimensions.

## 7.2. Clustering and querying of data

Utilizing the data embedding achieved through the VEOB and ST methods, the data can be organized into clusters using three distinct strategies:

1) The data can be divided into clustered cells as outlined in section 5, and the nearest cell can be searched to identify the most suitable matches.
2) The K-means algorithm can be employed to partition the data into cells, and again, the closest cell can be searched to find the best matches.
3) The embedding data can be loaded into a vector database, and the database's built-in nearest neighbor algorithm can be used to identify the best matches.

## 8. Apply VEOB and ST to supervised learning

Supervised learning aims to establish a mapping from an input vector $x$ to an output vector $y$, denoted as $x \rightarrow y$. This mapping can be represented as a probability of the eventual $y$ given an input $x$, or as a function $y = f(x)$. It includes both the training and prediction processes.

## 8.1. Training

Utilizing VEOB, the following steps are employed to train the mapping (for simplicity, the tilde bar on the symbol has been omitted):

1) The embeddings for $X$ and $Y$ are calculated using SVD. If $Y$ has a small number of categorical values, one-hot encoding is directly used as embedding, and the SVD step is not required:

$$P_x = U_x^T X \qquad P_y = U_y^T Y \ \ or \ \ P_y = Y$$

2) The $X$ and $Y$ embeddings are combined into a $Z$ matrix and its embedding is calculated as follows:

$$Z = \begin{bmatrix} P_x \\ \alpha P_y \end{bmatrix} \qquad P_z = U_z^T Z$$

Here, $\alpha$ is a scaling factor that ensures that the $X$ and $Y$ embeddings have similar weights.

3) The vectors of $P_z$ are divided into small cells. Each cell is a high-dimensional Voronoi polygon with a smooth relationship between x and y within it.
4) Within the cell, the probability of y occurrence can be computed, or a linear mapping between $P_x$ and $P_z$ can be used:

$$P_z = A_c P_x$$

Here, $A_c$ is a matrix which can be solved from matrices $P_x$ and $P_z$.

This method of using multiple cells for linear mapping is more akin to broad learning than deep learning. If a cell does not achieve the desired accuracy, it can be further subdivided into smaller cells.

## 8.2. Prediction

The matrices $U_x$, $U_y$, $U_z$ and $A_c$ obtained from the training process can be used for prediction. For a given input set $X$, the intermediate values can be calculated with the formular in the training steps 1), 2) and 3). The estimated $Z'$ can be calculated as follows:

$$Z' = U_z P_z$$

Select the $Z'$ value from the cell with the sub vectors $P'_x$ in $Z'$ that is closest to the $X$ embedding $P_x$. The predicted output can then be computed as:

$$Y' = U_y P'_y$$

In the case of y occurrence probability, the value with the highest probability in the cell can be used as the result.

## 9. Strategies for managing large data volumes

When dealing with a large dataset, such as billions of record lines, constructing the initial encoding matrix and performing the SVD can be technically challenging. The following two strategies are suggested as solutions.

### 9.1. Using incremental SVD

By utilizing the incremental SVD approach outlined in reference [12], the data vectors can be divided into smaller batches for easier handling. The incremental SVD is then applied repeatedly, ultimately resulting in the attainment of the global SVD at the final iteration step.

### 9.2. Partition the data into subdomains

The dataset is divided into smaller subsets that belong to different subdomains. In conjunction with vector database technology, these subdomains can be annotated with metadata, and each subdomain has its own vector embedding.

## 10. Conclusion

Harnessing the Power of VEOB and ST, we arrive at the following conclusions:

- Quick ML is realized through mathematically robust techniques that eliminate the need for slow back propagation steps.
- VEOB can reduce data dimensions, compress data, and generate a vector embedding that offers improved measurements of distances and similarities.
- ST can be utilized for long sequences of one-dimensional data like words and sentences, as well as two-dimensional data like images, benefiting from various tools from spectral analysis.
- The source data encoding preparation is straightforward, involving the creation of a column vector for each entity under consideration, followed by the construction of a matrix from these column vectors which is then fed into SVD for processing.
- The top SVD singular values offer a convenient method for selecting the vector embedding dimensions without significant information loss.
- For word to vector conversion, orthography embedding is highly stable and accurate. Semantic embedding is more reliable when using dictionary sources rather than context scanning of sample text corpora.

Areas for future exploration include:

- Expanding the application of VEOB to broader knowledge graph bases. The semantic embedding of words using a dictionary in our example represents a basic use case. More comprehensive research and experiments are needed on how to embed entities (nodes) and relationships (edges) in knowledge bases, especially when dealing with nodes and edges that possess many properties.
- VEOB and ST have the capacity to manage extremely long sequence data while simultaneously performing data compression. There is interest in exploring common use cases where transformer models are employed and enhancing their performance.

- Implementing VEOB and ST in the context of Reinforcement Learning (RL), given that RL involves a series of states, actions and rewards. ST could be utilized to transform the data into spectral space, and VEOB could be used for the data embedding of the states and actions of categorical data types.